\documentclass[12pt]{article}
\usepackage{graphicx}
\usepackage{epstopdf}
\usepackage{fancyhdr}
\usepackage{amsfonts}
\usepackage{amsmath}
\usepackage{fullpage}
\usepackage{upgreek}
\usepackage{float}
\usepackage{amsfonts}
\usepackage{enumitem}
\usepackage{caption}
\usepackage{subcaption}
\usepackage{url}
\usepackage{bm}
\usepackage{comment}
\usepackage{xcolor}

\usepackage{amsthm}
\newtheorem{result}{Result}

\usepackage{indentfirst}
\usepackage{multirow}
\usepackage{atbegshi}
\usepackage{afterpage}
\usepackage{natbib}
\usepackage{booktabs}
\usepackage{authblk}

\title{\bfseries{Robust XGBoosting for Regression}} 

\author[1]{Iris Aragón Mladosich}
\author[1]{Christophe Croux}

\affil[1]{ORSTAT, KU Leuven, Belgium}

\date{}

\begin{document}
	
	\maketitle
	
	\begin{abstract}
		XGBoost is a very popular and powerful method for prediction. It iteratively fits simple decision trees to the residuals of the previous step. An efficient and scalable implementation is available. The standard loss function for XGBoost is the quadratic loss, but a Huber loss can also be used. In this paper, we study the robustness of XGBoost and show that its performance can be affected by vertical outliers and leverage points. To address this, we explore alternative loss functions, based on M-, S-, and $\tau$-estimators from robust regression. Our results indicate that a two-step procedure, referred to as {MM-XGBoost}, provides the best trade-off between robustness and prediction accuracy. 
	\end{abstract}

	\section{Introduction}
	Machine learning has become an essential tool for predictive modeling. 
	Among the many available methods, gradient boosting \citep{friedman} is a particularly powerful approach, as it builds a strong predictor by iteratively combining simple base learners. At every iteration, a base learner is constructed by minimizing a loss function that measures the discrepancy between the model predictions and the observed response. Extreme gradient boosting, or XGBoost \citep{xgboost}, is an efficient and scalable implementation of gradient boosting. It is widely used due to its strong predictive performance, its built-in regularization, and its flexibility in handling different types of data. 
	
	In this paper, we show that the classical XGBoost for regression can be sensitive to outliers. Vertical outliers in the response variable and leverage points in the predictor variables can have a strong influence on the fitted model, reducing the accuracy and stability of the prediction. This sensitivity is mainly observed when the commonly used quadratic loss function of  XGBoost is used. To address this issue, the XGBoost implementation provides a more robust alternative in the form of a smooth approximation of the Huber loss. However, we show that even with this Huber loss, XGBoost can still be affected by leverage points.
	
	Motivated by these limitations, we aim to further improve the robustness of XGBoost for regression tasks by modifying the loss function used in the boosting procedure. In particular, we study several alternative loss functions, inspired by robust regression estimators. Implementations of XGBoost \citep{xgboostR} allow users to define custom loss functions without modifying the core algorithm, making it possible to incorporate such robust loss functions into the XGBoost algorithm without much difficulty.
	
	Our approach is based on ideas from \cite{friedman} and \cite{robustboost}, where the incorporation of robust loss functions into gradient boosting has been studied. In particular, \cite{friedman} and \cite{lutz} study loss functions based on M-estimators of regression using a robust scale estimate of the residuals. \cite{robustboost} use more sophisticated regression estimators and introduce loss functions based on S- and MM-estimators. We adapt and extend their ideas  to the extreme gradient boosting method. We also include the $\tau$-loss into the boosting procedure, which to the best of our knowledge was not considered before.
	
	Through simulation studies, we evaluate which of our proposed alternative loss functions improve robustness against outlying observations while still maintaining good predictive performance. For the remainder of the paper, Section~\ref{sec: xgb for regression} introduces the XGBoost method, Section~\ref{sec: methodology} introduces our proposed robust extensions of XGBoost, and Section~\ref{sec: sim studies} presents the simulation studies and discusses the results. Finally, Section~\ref{sec: conclusion} concludes the paper.

	\section{XGBoost for Regression} \label{sec: xgb for regression}
	
	In this section, we first outline the  principle of gradient boosting, followed by an introduction to extreme gradient boosting and the choice of loss functions. We adopt a vectorized notation for the training data. The number of predictor variables is denoted by $p$, and the number of observations by $n$.
	
	\subsection{Gradient Boosting}
	
	Consider a training dataset with covariate matrix  
	$\mathbf{X} = (\mathbf{x}_1, \dots, \mathbf{x}_n)^\top \in \mathbb{R}^{n \times p}$ and response vector  
	$\mathbf{y} = (y_1, \dots, y_n)^\top \in \mathbb{R}^n$. In gradient boosting with regression trees, the prediction model is constructed in a stage-wise manner by sequentially adding regression trees. Let $\hat{\mathbf{y}}^{(0)} \in \mathbb{R}^n$ denote the initial prediction, defined as a constant vector
	\[
	\hat{\mathbf{y}}^{(0)} = y^{(0)} \mathbf{1}_n,
	\]
	where $\mathbf{1}_n = (1 \dots 1)^\top \in \mathbb{R}$. Here, $y^{(0)} \in \mathbb{R}$ is an initial estimate of the response, typically chosen as the mean of $\mathbf{y}$.  
	
	{At each iteration, a regression tree
		\(f^{}:\mathbb{R}^p\rightarrow\mathbb{R}\) is fitted and added to the current model. A regression tree is defined by a tree structure
		\[
		q^{}:\mathbb{R}^p\rightarrow\{1,\dots,T\},
		\]
		which assigns each observation to one of the \(T\) leave nodes, and by an associated vector of leaf weights
		\[
		\mathbf w^{}=(w^{}_1,\dots,w^{}_{T})^\top .
		\]
		Accordingly, for an observation \(\mathbf x \in \mathbb{R}^p\), the tree prediction is given by the weight of the corresponding leaf node, i.e.,
		\[
		f^{}(\mathbf x)
		=
		w^{}_{q^{}(\mathbf x)}.
		\]
		Applying the tree to all observations in the training data yields the prediction vector
		\[
		F^{}(\mathbf X)
		=
		\big(
		f^{}(\mathbf x_1),
		\dots,
		f^{}(\mathbf x_n)
		\big)^\top
		\in \mathbb R^n.
		\]
		
		Let $f^{(k)}$ be the tree obtained in iteration $k$ with corresponding structure $q^{(k)}$ and associated weights $ \mathbf w^{(k)}$. 
		Let $\hat{\mathbf{y}}^{(k)} \in \mathbb{R}^n$ denote the vector of predictions of the $k$-th boosting iteration. At iteration $(k+1)$, the next regression tree is selected to improve the current model with respect to a specified loss function $\mathcal{L}$, which measures the discrepancy between the observed responses and the model predictions. More specifically, the tree $f^{(k+1)}$ is constructed such that, when added to the current fit, it  reduces the loss
		\begin{equation}
			\mathcal L\big(
			\mathbf y,
			\hat{\mathbf y}^{(k)} + F^{(k+1)}(\mathbf X)
			\big).
			\label{eq: loss simple}
		\end{equation}
		The resulting tree is added to the existing model, yielding
		\begin{equation}
			\hat{\mathbf y}^{(k+1)}
			=
			\hat{\mathbf y}^{(k)}
			+
			\eta F^{(k+1)}(\mathbf X),
			\label{eq: update}
		\end{equation}
		where \(\eta>0\) is a shrinkage parameter used to prevent overfitting.} The final predictions, after $K$ iterations, are given by
	\begin{equation*}
		\hat{\mathbf{y}}^{(K)}
		=
		\hat{\mathbf{y}}^{(0)} +  \eta \sum_{k=1}^K F^{(k)}(\mathbf{X}).
	\end{equation*}
	Thus, the learned predictor for a new observation $\mathbf{x} \in \mathbb{R}^p$ can be expressed as the ensemble of trees
	\begin{equation*}
		\hat{y}(\mathbf{x})
		=
		y^{(0)} +  \eta \sum_{k=1}^K f^{(k)}(\mathbf{x}),
	\end{equation*}
	where $f^{(k)}(\mathbf{x}) = w^{(k)}_{q^{(k)}(\mathbf{x})}$. We refer to Chapter 10 of \cite{elementsstatlearn} for more details on gradient boosting.
	
	\subsection{XGBoost}\label{subsec: xgboost}
	
	XGBoost is an extension of gradient boosting that aims to efficiently minimize the loss function \eqref{eq: loss simple} by using a second-order Taylor approximation. It also includes an explicit regularization term to control model complexity. At each boosting iteration, the method builds a regression tree based on this local approximation of the loss around the current predictions.
	
	More specifically, at iteration $k$, the loss is expanded around the prediction vector from the previous iteration, $\hat{\mathbf{y}}^{(k)}$, and a regularization term $\Omega(F^{})$ is added to {penalize the complexity of the newly added tree $F$}. Let $\mathbf{g}$ and $\mathbf{H}$ denote the gradient and Hessian of the loss function with respect to $\hat{\mathbf{y}}$, evaluated at $\hat{\mathbf{y}}^{(k)}$:
	\begin{equation}
		\mathbf{g}
		:= \nabla_{\hat{\mathbf{y}}} \mathcal{L}(\mathbf{y}, \hat{\mathbf{y}})
		\Big|_{\hat{\mathbf{y}} = \hat{\mathbf{y}}^{(k)}}
		\in \mathbb{R}^n,
		\qquad
		\mathbf{H}
		:= \nabla_{\hat{\mathbf{y}}}^{2} \mathcal{L}(\mathbf{y}, \hat{\mathbf{y}})
		\Big|_{\hat{\mathbf{y}} = \hat{\mathbf{y}}^{(k)}}
		\in \mathbb{R}^{n \times n}.
	\end{equation}
	The approximated loss function at iteration $k$ is then given by
	\begin{equation}\label{eq:taylor_loss}
		\mathcal{L} := \mathcal{L}\big(\mathbf{y}, \hat{\mathbf y}^{(k)} + F^{}(\mathbf{X})\big)
		\approx
		\mathbf{g}^\top F^{}(\mathbf{X})
		+
		\frac{1}{2} F^{}(\mathbf{X})^\top \mathbf{H} F^{}(\mathbf{X})
		+
		\Omega\big(F^{}\big),
	\end{equation}
	where the constant term $\mathcal{L}\big(\mathbf{y}, \hat{\mathbf{y}}^{(k)}\big)$ is omitted, as it does not influence the optimization.
	The regularization term $\Omega\big(F^{}\big)$ is given by
	\begin{equation*}
		\Omega\big(F^{}\big) = \gamma T + \frac{1}{2}\lambda \mathbf{w}^\top \mathbf{w},
	\end{equation*}
	where $T$ denotes the number of leaves in the tree and $\mathbf{w}$ the corresponding leaf weights. The parameters $\gamma$ and $\lambda$ control the penalization of $T$ and $\mathbf{w}$, respectively. 
	
	For a fixed tree structure \(q\), let
	\(S\in\{0,1\}^{n\times T}\) denote the leaf-assignment matrix, defined by
	\[
	S_{ij}
	=
	\begin{cases}
		1, & \text{if } q(\mathbf x_i)=j,\\
		0, & \text{otherwise}.
	\end{cases}
	\]
	Each row of \(S\) contains exactly one nonzero entry, indicating the leaf to which the corresponding observation of the training data  is assigned. 
	The vector of tree predictions can be now be written as
	\[
	F^{}(\mathbf X)=S\mathbf w.
	\]
	Thus, once the structure $q$ is fixed, determining the tree reduces to finding the optimal the leaf weights \(\mathbf w\).{Substituting \(F^{}(\mathbf X)=S\mathbf w\) into the approximation
		\eqref{eq:taylor_loss} yields
		\begin{equation*}\label{eq: approx loss}
			\mathcal L
			\approx
			\mathbf g^\top S\mathbf w
			+
			\frac12 (S\mathbf w)^\top \mathbf H (S\mathbf w)
			+
			\frac12 \lambda \mathbf w^\top \mathbf w
			+
			\gamma T.
		\end{equation*}
		Differentiating with respect to \(\mathbf w\) and setting the derivative equal to zero gives the optimal leaf weights
		\begin{equation}\label{eq: optimal weights}
			\mathbf w^* = -\big(S^\top \mathbf H S+\lambda I\big)^{-1}S^\top\mathbf g.
		\end{equation}
		Hence, for a given tree structure, the optimal leaf weights can be obtained directly from the first- and second-order derivatives of the loss function. The tree structure $q$ is constructed greedily in XGBoost by recursively evaluating candidate splits using a split-gain criterion derived from the regularized objective. 
		The process continues until no further split provides sufficient improvement or a predefined stopping criterion is reached.}
	
	\paragraph{Built-in loss functions.}
	The original XGBoost implementation of \cite{xgboost} assumes the loss to be additive over individual observations. The loss can be expressed in terms of the residuals $r_i = y_i - \hat{y}_i$, for $i = 1, \dots,n$, as
	\begin{equation}\label{eq: loss rho}
		\mathcal{L}(\mathbf{y}, \hat{\mathbf y}) = \sum_{i=1}^n \rho(r_i),
	\end{equation}
	where $\rho: \mathbb{R} \rightarrow \mathbb{R}^+$ is an even positive function.
	This $\rho$-function determines how prediction errors are penalized and therefore controls robustness. 
	
	The default choice in the XGBoost implementation \citep{xgboostR} for $\rho$ is the quadratic $\rho$-function
	\begin{equation*}\label{eq: rho quad}
		\rho(u) = \tfrac{1}{2}u^2.
	\end{equation*}
	This $\rho$-function is optimal under Gaussian noise, but is highly sensitive to outliers as it penalizes large residuals quadratically. {We will refer to XGBoost using the quadratic loss as XGB Quadratic.} In Section~\ref{sec: sim studies} we show that XGB Quadratic is sensitive to both vertical outliers and leverage points in the training data.
	
	To improve robustness, the XGBoost implementation provides an alternative  choice, the Pseudo-Huber $\rho$-function \citep{PseudoHuber_Charbonnier, PseudoHuber_Zhang} 
	\begin{equation*}\label{eq: rho pseudo huber}
		\rho(u) 
		= 
		b^2\big( \sqrt{1 + \big(\frac{u}{b}\big)^2} - 1 \big),
	\end{equation*}
	which is a smooth (twice differentiable) alternative for the Huber $\rho$-function \citep{M-est}
	\begin{equation} \label{eq: rho huber}
		\rho(u) =
		\begin{cases}
			\frac{1}{2}u^2, & |u| \leq b, \\
			b\big(|u| - \frac{b}{2}\big), & |u| > b.
		\end{cases}
	\end{equation}
	The tuning parameter $b>0$ controls the transition between its quadratic and linear behavior. Small residuals are penalized quadratically, whereas large residuals are penalized only linearly to reduce their influence. In the XGBoost implementation \citep{xgboostR}, the default tuning constant of the Pseudo-Huber $\rho$-function is $b = 1$. We refer to XGBoost using the Pseudo-Huber loss as XGB Pseudo-Huber. 
	
\begin{figure}
		\centering
		\includegraphics[width=0.9\linewidth]{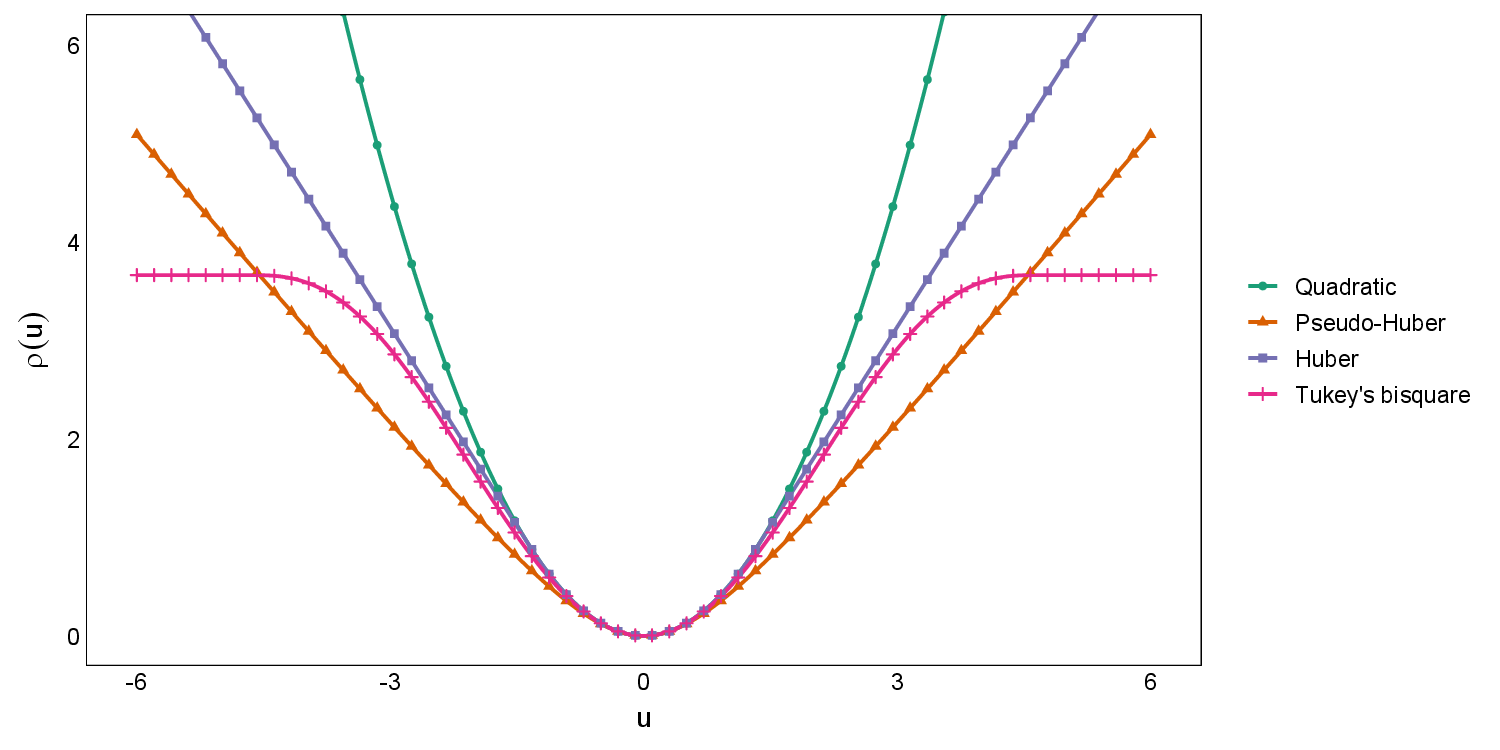}
		\caption{Comparison of different $\rho$-functions: Quadratic, Pseudo-Huber (with $b=1$), Huber (with $b=1.345$), and Tukey's bisquare (with $c=4.865$).}
		\label{fig: rho functions}
	\end{figure}

	Because XGB Pseudo-Huber penalizes large residuals linearly, unlike XGB Quadratic, it is expected to be more robust to outliers. In the simulation studies presented in Section~\ref{sec: sim studies}, we confirm this improvement in the presence of vertical outliers. However, the results also show that its performance is still highly sensitive to leverage points. This motivates the need for incorporating alternative loss functions within the boosting framework that are more robust than the loss functions provided by the standard XGBoost implementation. An alternative loss function, very popular in robust statistics, is the  Tukey's bisquare function \citep{Tukey}
	\begin{equation} \label{eq: rho tukey}
		\rho(u) =
		\begin{cases}
			\tfrac{c^2}{6}\big[1 - \big(1 - \big(\dfrac{u}{c}\big)^2\big)^3\big], & |u| \leq c, \\
			\frac{c^2}{6}, & |u| > c.
		\end{cases}
	\end{equation}
	This $\rho$-function flattens out beyond the cutoff $c>0$, such that residuals with absolute value larger than $c$ have constant influence on the fit. The derivative $\psi$ of the Tukey's bisquare $\rho$-function is redescending to zero.
	The tuning constant $c$ controls the trade-off between robustness and efficiency: a smaller value of $c$ yields higher robustness by downweighting moderate residuals, while larger values allow more observations to contribute fully  to the model fit.  
	The selected constant $c$ ensures high efficiency (of $95\%$) under Gaussian errors while still providing high robustness to outliers \citep{Maronna_Robust_Book}. 
	Figure~\ref{fig: rho functions} plots the considered $\rho$-functions. 

	\section{Robust XGBoost}\label{sec: methodology}

	We propose several variants of XGBoost, each defined by a loss function inspired by well-known robust regression estimators. 
	Specifically, we introduce \textit{M-XGBoost}, based on M-estimators \citep{M-est}, \textit{S-XGBoost}, based on S-estimators \citep{S-est}, and {$\tau^2$}\textit{-XGBoost}, based on $\tau$-estimators \citep{tau-est}. In addition, we consider a two-stage procedure, \textit{MM-XGBoost}, similar in spirit to MM-estimators \citep{MM-est}. The latter combines the high robustness of S-estimators with the more efficient M-estimators. We refer to Chapter $4$ and $5$ of \cite{Maronna_Robust_Book} for an overview of robust regression methods.
	
	The XGBoost method requires the gradient $\mathbf{g}$ and Hessian $\mathbf{H}$ in the second-order Taylor expansion of the loss function~\eqref{eq:taylor_loss}. Note that for an additive loss function of the form \eqref{eq: loss simple}, all cross-partial derivatives vanish, yielding a diagonal Hessian matrix $\mathbf{H}$. Since this Hessian matrix $\mathbf{H}$ becomes intractably large for $n$ big, we approximate it by its diagonal. We denote the diagonal vector of $\mathbf{H}$ by
	\begin{equation*} \label{eq: hess vec}
		\mathbf{h}=  \big(
		\frac{\partial^2 \mathcal{L}}{\partial \hat y_1^2},
		\dots,
		\frac{\partial^2 \mathcal{L}}{\partial \hat y_n^2}
		\big)^\top.
	\end{equation*}

	So, to implement our custom loss functions, we only need to derive the corresponding first- and second-order derivatives
	\begin{equation}\label{eq: grad and hess entries}
		g_j=\frac{\partial \mathcal{L}(\mathbf{y}, \hat{\mathbf{y}})}{\partial \hat y_j}
		\quad \text{and} \quad
		h_j = \frac{\partial^2 \mathcal{L}(\mathbf{y}, \hat{\mathbf{y}})}{\partial \hat y_j^2},
	\end{equation}
	for $j = 1,\dots,n.$
	In the next subsections we explicitly derive these quantities for each of the proposed XGBoost methods.

	\subsection{M-XGBoost}\label{sec: m-xgboost}
	
	The first method we consider is the M-XGBoost method. Its loss function is similar to \eqref{eq: loss rho}, but uses scaled residuals. The resulting M-loss is defined as
	\begin{equation}\label{eq: M-XGBoost loss}
		\mathcal{L}_M(\mathbf{y}, \hat{\mathbf{y}})
		= \sum_{i=1}^n \rho\big(\frac{r_i}{\sigma_0}\big)\sigma_0^2,
	\end{equation}
	where $r_i = y_i - \hat{y}_i$ denotes the residual of the $i$-th observation, for $i = 1, \dots,n$. The scale $\sigma_0$ is treated as a fixed robust scale estimate, computed once from the response variable before the boosting iterations begin. For M-XGBoost, we use the median absolute deviation (MAD) of the responses
	\begin{equation*}\label{eq: mad scale}
		\sigma_0 = 1.48\cdot\operatorname{med}_i\, | y_i - \operatorname{med}_j\, y_j |,
	\end{equation*}
	a well-known robust alternative to the standard deviation \citep{Maronna_Robust_Book}. The $\rho$-function in \eqref{eq: M-XGBoost loss} is the Huber function \eqref{eq: rho huber} with tuning constant $b = 1.345$, discussed in Section~\ref{subsec: xgboost}. Note that setting $\sigma_0 = 1$ reduces the M-loss in \eqref{eq: M-XGBoost loss} to the standard XGBoost loss in \eqref{eq: loss rho}. Furthermore, choosing the quadratic $\rho$-function reduces the M-loss to the classical squared error loss.
	
	To incorporate the M-loss~\eqref{eq: M-XGBoost loss} into the XGBoost framework, we derive the corresponding gradient and diagonal Hessian entries in Equation~\eqref{eq: grad and hess entries}. 
	Since the scale $\sigma_0$ is treated as fixed and computed in advance from the response variable, the M-loss depends on $\hat{\mathbf y}$ only through the residuals. Throughout the paper, we denote by $\psi$ and $\psi'$ the first and second derivatives of the $\rho$-function. We obtain the following result (the proof can be found in \ref{appen: sec: proof m-loss}).
	
	\begin{result}\label{result: grad and hess M}
		The components of the gradient $\mathbf{g}$ and diagonal Hessian $\mathbf{h}$ of the M-loss \eqref{eq: M-XGBoost loss} are
		\begin{equation}\label{eq:grad m-xg}
			g_j
			= - \psi(\tilde{r}_{j,0})\, \sigma_0 \  \mbox{\  and \ } \  h_j
			= -\psi'(\tilde{r}_{j,0})
		\end{equation}
		where $\tilde{r}_{j,0} = {{r}_{j}}/{\sigma_0}$ for $j =1, \dots, n$.
	\end{result}

	\subsection{S-XGBoost}\label{sec: s-xgboost}
	
	In the M-loss, the scale $\sigma_0$ is treated as known and does not depend on the regression fit. Although such loss functions can provide some robustness against outliers, it is better to estimate the scale jointly with the model predictions. To address this, we consider the S-XGBoost method, where the loss function is
	\begin{equation}\label{eq: s-loss}
		\mathcal{L}_S = \tfrac{1}{2} n \sigma^2,
	\end{equation}
	with $\sigma$ a robust estimator of the residual scale. For S-XGBoost, $\sigma$ is taken as an M-estimator of scale centered at zero, implicitly defined through the M-scale equation \citep{M-est}
	\begin{equation}\label{eq:m-scale sxg}
		\frac{1}{n}\sum_{i=1}^n \rho_1\big(\frac{r_i}{\sigma}\big) = \kappa.
	\end{equation} 
	The constant $\kappa$ is taken as $\kappa = \mathbb{E}[\rho_1(\varepsilon)]$ for $\varepsilon \sim \mathcal{N}(0,1)$ \citep{Maronna_Robust_Book}. We take $\rho_1$ as Tukey’s bisquare $\rho$-function~\eqref{eq: rho tukey}. Choosing the tuning constant $c = 1.548$ yields a 50\% breakdown point and corresponds to $\kappa = 0.2$. 
	
	To incorporate the S-loss~\eqref{eq: s-loss} into the XGBoost framework, we need to derive the corresponding gradient and diagonal Hessian entries in Equation~\eqref{eq: grad and hess entries}. The $\sigma$ in \eqref{eq: s-loss} is not fixed, but is implicitly defined through the M-scale~\eqref{eq:m-scale sxg}. Since it depends on the residuals $r_i = y_i - \hat{y}_i$, it therefore depends on the prediction vector $\hat{\mathbf y}$. We obtain the following result (the proof can be found in \ref{appen: sec: proof s-loss}).

	\begin{result}\label{result: grad and hess S}
		The components of the gradient $\mathbf{g}$ and diagonal Hessian $\mathbf{h}$ of the S-loss \eqref{eq: s-loss}, for $j=1,\dots,n$, are
		\begin{equation}\label{eq:gradS}
			g_j
			= -\,n \sigma\,\frac{\psi_1(\tilde{r}_j)}{A_1}
		\end{equation}
		and
		\begin{equation}\label{eq:hessS}
			h_j= n\big(
			\frac{\psi_1'(\tilde{r}_j)}{A_1}
			- 2\,\frac{\psi_1'(\tilde{r}_j)\psi_1(\tilde{r}_j)\tilde{r}_j}{A_1^2}
			+ \frac{\psi_1^2(\tilde{r}_j)}{A_1^2}\big(\frac{B_1}{A_1} - 1\big)
			+ \frac{\psi_1^2(\tilde{r}_j)}{A_1^2}
			\big),
		\end{equation}
		where $A_1 = \sum_{i=1}^n \tilde{r}_i\,\psi_1(\tilde{r}_i)$ and $B_1 = \sum_{i=1}^n 
		\tilde{r}_i\big(\psi_1(\tilde{r}_i)+\tilde{r}_i\psi_1'(\tilde{r}_i)\big)$, with $\tilde r_i = r_i/\sigma$ for $1 \leq i \leq n$.
	\end{result}

	Note that for the quadratic $\rho$-function with $\kappa = {1}/{2}$, the S-loss reduces to the classical squared-error loss, since $\sigma^2 = \sum_{i=1}^n r_i^2/n$. Consequently, $A_1 = \sum_{i=1}^n \tilde{r}_i^2$ and $B_1 = 2A_1$, which implies $g_j = -r_j$ and $h_j = 1$ for $j = 1, \dots, n$. These derivatives coincide with Result~\ref{result: grad and hess M}.

	S-estimators of scale offer strong robustness, but have reduced efficiency under clean data. Therefore, S-XGBoost is expected to achieve high robustness, but possibly larger values for the prediction errors, in particular when no outliers are present.  To address this issue, we extend the ideas of $\tau$- and MM-estimators \citep{Maronna_Robust_Book} to XGBoost, resulting in $\tau^2$-XGBoost and MM-XGBoost. The idea is that the improved efficiency of the regression estimators will translate into  higher prediction accuracy under clean and contaminated data.
	
	\subsection{\boldsymbol{$\tau^2$}-XGBoost}\label{sec: tau-xgb}
	Like S-XGBoost, $\tau^2$-XGBoost minimizes a robust residual scale, 
	but replaces the M-scale by the  more efficient $\tau$-scale. The $\tau^2$-loss function is defined as
	\begin{equation}\label{eq: tau2 loss}
		\mathcal{L}_{\tau^2} = \frac{n}{2\kappa}\,\tau^2,
	\end{equation}
	where the $\tau$-estimator of scale is defined as
	\begin{equation}\label{eq:tau2-def}
		\tau^2 
		= \frac{1}{n} \sum_{i=1}^n 
		\rho\big(\frac{{r}_i}{\sigma}\big)\sigma^2,
	\end{equation}
	with residuals $r_i = y_i - \hat{y}_i$  \citep{tau-est} .  
	The scale $\sigma$ is a M-scale, defined in Equation~\eqref{eq:m-scale sxg}, where the function $\rho_1$ is the Tukey’s bisquare $\rho$-function with $c_1 = 1.548$. 
	For the $\rho$-function in \eqref{eq:tau2-def}, we take the Tukey’s bisquare $\rho$-function, now with a lager value for the tuning constant, namely $c = 4.865$, to ensure high efficiency under Gaussian errors while maintaining strong robustness.

	To incorporate the $\tau^2$-loss~\eqref{eq: tau2 loss} into the XGBoost framework, we derive the corresponding gradient and diagonal Hessian entries in Equation~\eqref{eq: grad and hess entries}.
	Note that $\tau^2$, as defined in \eqref{eq:tau2-def}, depends on the residuals both directly and through the scale $\sigma$. 
	We obtain the following result (the proof can be found in \ref{appen: sec: proof tau-loss}).
	
	\begin{result} \label{result: grad and hess tau}
		The components of the gradient $\mathbf{g}$ and diagonal Hessian $\mathbf{h}$ of the $\tau^2$-loss \eqref{eq: tau2 loss}, for $j=1,\dots,n$, are
		\begin{equation}\label{eq:grad-tau2}
			g_j
			= -
			\frac{\psi_1(\tilde{r}_j)\sigma}{2\kappa}
			\big(\frac{2C}{A_1}-\frac{A}{A_1}\big)
			- \frac{\psi(\tilde{r}_j)\sigma}{2\kappa}
		\end{equation}
		and
		\begin{align}\label{eq:final_Hessian_tau2}
			h_j
			= \frac{1}{2\kappa}\Big[
			&\big(\frac{\partial\sigma}{\partial\hat y_j}\big)^2 (2C + B - 3A) \nonumber
			+ \big(\frac{\partial^2\sigma}{\partial\hat y_j^2}\big)\big(2\sigma C - \sigma A\big) \nonumber\\
			+ &\big(\frac{\partial\sigma}{\partial\hat y_j}\big)\big(2 \psi'(\tilde{r}_j)\tilde{r}_j - 2\psi(\tilde{r}_j)\big)
			+ \psi'(\tilde{r}_j)
			\Big],
		\end{align}
		where
		$
		A = \sum_{i=1}^n \tilde{r}_i \psi(\tilde{r}_i), 
		C = \sum_{i=1}^n \rho(\tilde{r}_i)$, $
		A_1 = \sum_{i=1}^n \tilde{r}_i \psi_1(\tilde{r}_i)
		$ and
		$B = \sum_{i=1}^n \tilde{r}_i\big(\psi(\tilde{r}_i) + \tilde{r}_i\psi'(\tilde{r}_i)\big)$, with $\tilde r_i = r_i/\sigma$ for $1 \leq i \leq n$.
	\end{result}

	For $\rho = \rho_1$ with equal tuning constants, the $\tau^2$-loss coincides with the S-loss. {Then} $\psi = \psi_1$, $A = A_1$ and $C = n\kappa$. The $j$-th gradient entry of the $\tau^2$-loss, for $j=1,\dots,n$, simplifies to
	$g_j
	= -n\sigma\,{\psi(\tilde{r}_j)}/{A}$, which is exactly the S-loss gradient entry~\eqref{eq:gradS}. An analogous simplification holds for the diagonal Hessian entry.
	
	\subsection{MM-XGBoost}\label{sec: mm-xgboost}
	
	In addition to $\tau$-estimators, MM-estimators provide another approach to improve efficiency while maintaining strong robustness properties \citep{MM-est}. We adapt this idea to the XGBoost framework and refer to the resulting method as MM-XGBoost. The method follows a two-stage procedure that combines a highly robust initial fit with a refinement step. In the first stage, we obtain robust predictions using S-XGBoost. In the second stage, a robust scale estimate is computed from the residuals of the initial fit. The predictions from the first stage are used as the starting vector in the second stage.
	In summary,
	\paragraph{Stage 1: Robust Initial Fit.}  
	Apply S-XGBoost (Section~\ref{sec: s-xgboost}) with $K_1$ boosting iterations to obtain a highly robust initial predictor. Let
	\[
	r_{S,i} = y_i - \hat{y}_{S,i}, \qquad i = 1,\dots,n,
	\]
	denote the residuals from the final S-XGBoost prediction. Define the scale parameter $\sigma_0$ as the M-estimator of scale \eqref{eq:m-scale sxg} computed from the residuals $r_{S,i}$, with $\rho_1$ the Tukey’s bisquare function \eqref{eq: rho tukey} with tuning constant $c = 1.345$. 
	
	\paragraph{Stage 2: Refinement.}  
	Apply XGBoost with the M-loss defined in Equation~\eqref{eq: M-XGBoost loss} with $K_2$ boosting iterations, using as initial predictions the final output from stage~1:
	\[
	\hat{\mathbf y}^{(0)} = (\hat y_{S,1}, \dots, \hat y_{S,n})^\top.
	\]
	In this second stage, the M-loss employs the Tukey’s bisquare $\rho$-function \eqref{eq: rho tukey} with tuning constant $c = 4.685$, which ensures high efficiency under Gaussian errors while maintaining robustness. The scale $\sigma_0$ is the M-estimator of scale obtained in stage~1. We take the number of boosting iterations $K_1$ in the first step  equal to the number of boosting iterations $K_2$ in the second step.
	
	\section{Simulation Study}\label{sec: sim studies}
	
	To evaluate the robustness and predictive performance of the proposed XGBoost-based methods introduced in Section~\ref{sec: methodology}, we conducted a simulation study. Robustness is assessed by comparing prediction performance when models are trained under various contamination scenarios. In particular, we compare the results obtained from the contaminated training data with those in the clean data case. We consider the baseline methods {XGB Quadratic} and {XGB Pseudo-Huber} together with the proposed XGBoost variants: M-XGBoost, S-XGBoost, \linebreak $\tau^2$-XGBoost, and MM-XGBoost. After presenting the  simulation settings, we discuss the details of the model implementation. Next, we present the performance measure used and we discuss the simulation results at length.
	\subsection{Data Generation}
	Consider a training set $\{(\mathbf{x}_i, y_i), i = 1, \dots, n\}$ consisting of predictor variables $\mathbf{x}_i \in \mathbb{R}^p$ and responses $y_i \in \mathbb{R}$. The responses are generated from the regression model
	\begin{equation*}
		y_i = g(\mathbf{x}_i) + \varepsilon_i, \qquad i=1, \dots,n,
	\end{equation*}
	where the errors $\varepsilon_i$ are $\text{i.i.d. } \mathcal{N}(0,1)$ and independent of the predictor variables $\mathbf{x}_i$. We consider several regression functions for $g(\cdot)$. Let $p \geq 5$ and write $\mathbf{x} = (x_1, \dots, x_p)^\top$.  We consider a low-dimensional setting with $p=10$ covariates, and a high-dimensional setting with $p=100$. The first regression function is the linear function $g_1(\cdot)$:
	\begin{equation*}
		g_1(\mathbf{x}) = \sum_{j=1}^5 x_j.
	\end{equation*}
	In addition, we include two non-linear regression functions $ g_2(\cdot)$ and $g_3(\cdot)$ adapted from the simulation settings in \cite{robustboost}:

	\begin{align*}
		g_2(\mathbf{x}) 
		&= 5 \sqrt{ {x}_{1}^2 
			+ \big( {x}_{2} {x}_{3} 
			- \frac{1}{{x}_{2} {x}_{4}} \big)^2 }, \\[1mm]
		g_3(\mathbf{x}) 
		&= 
		\big[ \sum_{j=1}^5 
		\big( 1 + (-1)^j 0.8 {x}_{j} 
		+ \sin(6{x}_{j}) \big) \big]
		\big[ \sum_{j=1}^3 
		\big( 1 + \frac{{x}_{j}}{3} \big) \big].
	\end{align*}
	
	The components of the predictor variables are generated from a uniform distribution $\mathcal{U}(0,1)$, except for $g_2$, where $x_{2}$ and $x_{4}$ follow a $\mathcal{U}(1,2)$ distribution due to constraints on $g_2$.
	
	For the correlation structures of the predictor variables, we consider three scenario's, again adapted from the simulation settings of \cite{robustboost}. The first structure, $S_1$, assumes independent components. So the correlation matrix of $\mathbf{x}$, $\mathrm{Cor}(\mathbf{x})$, is diagonal. 
	The second structure, $S_2$, introduces a decreasing correlation pattern
	\[
	\mathrm{Cor}(\mathbf{x})_{jk} = 0.8^{|j-k|}.
	\]
	This structure induces stronger dependence between predictors with nearby indices and weaker dependence as their indices are farther apart. The third structure, $S_3$, follows a block correlation pattern
	\[
	\mathrm{Cor}(\mathbf{x})_{jk} =
	\begin{cases}
		0.8 & \text{if } x_{j} \text{ and } x_{k} \text{ belong to the same block}, \\[1mm]
		0   & \text{otherwise}.
	\end{cases}
	\]
	The composition of the blocks depends on the underlying regression function. For $g_1$ and $g_3$, block 1 consists of $(x_{1}, x_{2}, x_{3})$ and block 2 of $(x_{4}, x_{5})$. For $g_2$, the blocks are $(x_{1}, x_{2})$ and $(x_{3}, x_{4})$. This structure represents settings in which predictors are strongly correlated within groups but independent between groups.
	
	We consider two contamination rates in the training data: low contamination ($5\%$) and high contamination ($20\%$). Let $\mathcal{I}_{\text{out}} \subseteq \{1,\dots,n\}$ denote a randomly selected index set of contaminated observations. We consider the following three contamination scenarios along with the clean data case.
	
	\paragraph{(a) Vertical Outliers.}
	The responses are shifted:
	\[
	y_i^{(\text{vertical})} = y_i + \delta_i^y, \qquad \text{for}\quad i \in \mathcal{I}_{\text{out}},
	\]
	where $\delta_i^y \sim t_1$, a Student-$t$ distribution with 1 degree of freedom. The predictor variables remain unchanged.
	
	\paragraph{(b) Good Leverage Points.}
	The predictor variables are shifted:
	\[
	{x}_{ij}^{(\text{good})} = {x}_{ij} + \delta_{ij}^x, 
	\qquad \text{for}\quad i \in \mathcal{I}_{\text{out}},
	\]
	where $\delta_{ij}^x \sim \mathcal{N}(\mu_{ij}, 10)$ and $\mu_{ij} \sim \mathcal{N}(10,10)$, for $j = 1,\dots,p$. 
	The response values are then regenerated according to the model:
	\[
	y_i^{(\text{good})} = g\big(\mathbf{x}_i^{(\text{good})}\big) + \varepsilon_i,
	\qquad \text{for}\quad i \in \mathcal{I}_{\text{out}}.
	\]
	
	\paragraph{(c) Bad Leverage Points.}
	First, predictor variables are shifted:
	\[
	{x}_{ij}^{(\text{bad})} = {x}_{ij} + \delta_{ij}^x, 
	\qquad \text{for}\quad i \in \mathcal{I}_{\text{out}},
	\]
	where $\delta_{ij}^x \sim \mathcal{N}(\mu_{ij}, 10)$ and $\mu_{ij} \sim \mathcal{N}(10,10)$, for $j = 1,\dots,p$. 
	Then, the responses are regenerated and additionally shifted:
	\[
	y_i^{(\text{bad})}
	= g\big(\mathbf{x}_i^{(\text{bad})}\big) + \varepsilon_i + \delta_i^y,
	\qquad \text{for}\quad i \in \mathcal{I}_{\text{out}},
	\]
	where $\delta_i^y \sim t_1$.

	\subsection{Model Implementation} \label{subsection:model}
	The loss functions for the baseline methods (XGB Quadratic and XGB Pseudo-Huber) are built-in options of \texttt{xgboost} in R/Python. In contrast, the  other methods (M-XGBoost, S-XGBoost, $\tau^2$-XGBoost, and MM-XGBoost)  use custom loss functions,  which are supported by \texttt{xgb.train}. Such a custom loss function requires the vectors $\hat{\mathbf y}$ and ${\mathbf y}$ as inputs, and should return the  gradient vector $\mathbf g$ 
	and the diagonal Hessian vector $\mathbf h$ as output. Hence, it is not necessary to compute the loss function $\mathcal L$ itself, as one only needs $\mathbf g$ and $\mathbf h$ for the optimal leaf weights in  \eqref{eq: optimal weights}. Implementing the custom loss functions therefore becomes immediate, as one only needs to code  the explicit expressions for  $\mathbf g$ and $\mathbf h$ given in Results 1-3 of  Section~\ref{sec: methodology}. 
	
	The initial prediction used as a start for the boosting iterations is chosen as the median of the response values
	\[
	{y}^{(0)} 
	= \operatorname{median}(y_1,\dots,y_n).
	\]
	This provides a more robust starting point than the commonly used mean, and we use it for all methods in this study. As will be shown, only taking a robust starting value is not sufficient to obtain a fully robust version of XGBoost. The number of boosting iterations is fixed at $K = 100$. For larger values of $K$ no further improvement in prediction accuracy was observed, for the simulation settings considered.  We preferred to work with a fixed choice for the number of iterations, to focus solely on the  comparison  of the different loss functions. An alternative strategy is to select $K$ using a validation set, combined with an early stopping rule. 
	For MM-XGBoost we set {$K_1 = 50$} iterations in stage~1 and {$K_2 = 50$} iterations in stage~2, resulting in the same total number of iterations as for the other XGBoost methods. Numerical experiments indicate that choosing $K_2 \geq K_1$  yields the best performance. The first stage is designed to ensure robustness and should therefore not involve an excessive number of iterations, whereas the second stage focuses on improving accuracy and thus benefits from a larger number of iterations.

	A set of tuning parameters must be specified when applying XGBoost.  The maximum tree depth is fixed at $d = 1$, thereby restricting the base learners to decision stumps. 
	
	The tree construction method is set to \texttt{exact},  which employs an exact greedy search for optimal split finding rather than a histogram based method \citep{xgboost}. All other hyperparameters of XGBoost in the R/Python implementation \citep{xgboostR} are kept at their default. These include the learning rate $\eta$ (default value $0.3$), the regularization parameters $\gamma$ and $\lambda$ (default values $0$ and $1$, respectively), and the minimum sum of instance weights required in a leaf node (default value $1$).

	\subsection{Performance Measure}
	
	For each simulation run, we randomly partition the data into a training set, used to fit the model, and a test set, used exclusively to evaluate predictive performance. Predictive performance is evaluated using the Root Mean Squared Error (RMSE).
	The test set is kept clean, which means that no vertical outliers nor leverage points are introduced into it.  Let $\mathcal{I}_{\text{test}}$ denote the index set of test observations and $\hat{y}_i$ the prediction for observation $i \in \mathcal{I}_{\text{test}}$. The RMSE is defined as
	\begin{equation*}
		\text{RMSE} 
		= \sqrt{
			\frac{1}{|\mathcal{I}_{\text{test}}|}
			\sum_{i \in \mathcal{I}_{\text{test}}}
			\big(y_i - \hat y_i\big)^2
		}.
	\end{equation*}
	
	Each simulation scenario is replicated $100$ times to ensure stable performance estimates. For every scenario, we report the mean RMSE across the replicates and its corresponding standard error (SE). An alternative performance measure is the Mean Absolute Error, yielding results that are fully in line with the ones presented.

	\subsection{Results}
	
	For each combination of regression function, dimensionality, correlation structure, and contamination scenario, we evaluate the predictive performance of all proposed XGBoost variants as well as the baseline methods XGB Quadratic and XGB Pseudo-Huber. The training sample size is fixed at $n_{\text{train}} = 700$ and the test sample size at $n_{\text{test}} = 300$.  We discuss two settings in detail. The tables of the RMSE for all other simulation settings can be found in the supplementary material. 
	
	\paragraph{Setting 1: Non-linear function, low-dimensional, correlated predictors, low contamination rate.}
	
	The first setting considers the non-linear regression function $g_2$ in a scenario with $p=10$ predictors following correlation structure $S_3$, and a contamination rate of $5\%$. Figure~\ref{fig: boxplot g2, S2, 0.05} presents the boxplots of the $\log(\text{RMSE})$ over $100$ simulation replicates. Table~\ref{tab:g2_S2_005 text} reports the mean RMSE values with standard errors. As expected, XGB Quadratic is not robust. It is affected by both vertical outliers and leverage points, resulting in increased prediction error and variability compared to the clean data case. Applying XGBoost with the more robust Pseudo-Huber loss function improves performance in the presence of vertical outliers, but remains sensitive to leverage points. A similar pattern is observed for M-XGBoost: Although it provides robustness against vertical outliers, it remains highly sensitive to leverage points. In contrast, S-XGBoost shows strong robustness across all contamination scenarios, with stable prediction errors. However, this robustness comes at the cost of slightly reduced efficiency in the clean data case, as it yields higher RMSE values than the baseline methods. The $\tau^2$-XGBoost method improves efficiency relative to S-XGBoost, achieving lower RMSE in the clean data case while maintaining good robustness.
	Finally, MM-XGBoost balances robustness and efficiency most effectively. It substantially improves prediction error compared to S-XGBoost, more so than $\tau^2$-XGBoost, while preserving a high level of robustness. Across all contamination scenarios, it achieves overall the lowest RMSE with low variability, making it the overall best-performing method in this setting.

	\begin{table}
		\centering
		\caption{\label{tab:g2_S2_005 text}Mean RMSE (SE) for $g_2$ with $p=10$ under clean and 5\% contamination scenarios with $S_3$ for $M = 100$ independent runs}
		\centering
		\begin{tabular}[t]{lcccc}
			\toprule
			Model & clean & vertical & good leverage & bad leverage\\
			\midrule
			XGB Quadratic & 1.27 (0.01) & 1.96 (0.32) & 2.23 (0.02) & 2.58 (0.24)\\
			XGB Pseudo-Huber & 1.28 (0.01) & 1.29 (0.01) & 4.91 (0.13) & 5.45 (0.15)\\
			\midrule
			M-XGBoost & 1.28 (0.01) & 1.29 (0.01) & 9.08 (0.34) & 10.03 (0.35)\\
			\midrule
			S-XGBoost & 1.57 (0.01) & 1.58 (0.01) & 1.54 (0.01) & 1.56 (0.01)\\
			\midrule
			$\tau^2$-XGBoost & 1.31 (0.01) & 1.33 (0.01) & 1.37 (0.01) & 1.38 (0.01)\\
			\midrule
			\addlinespace
			MM-XGBoost & 1.28 (0.00) & 1.28 (0.01) & 1.28 (0.01) & 1.29 (0.01)\\
			\bottomrule
		\end{tabular}
	\end{table}
	
	\begin{figure}
		\centering
		\includegraphics[width=\linewidth]{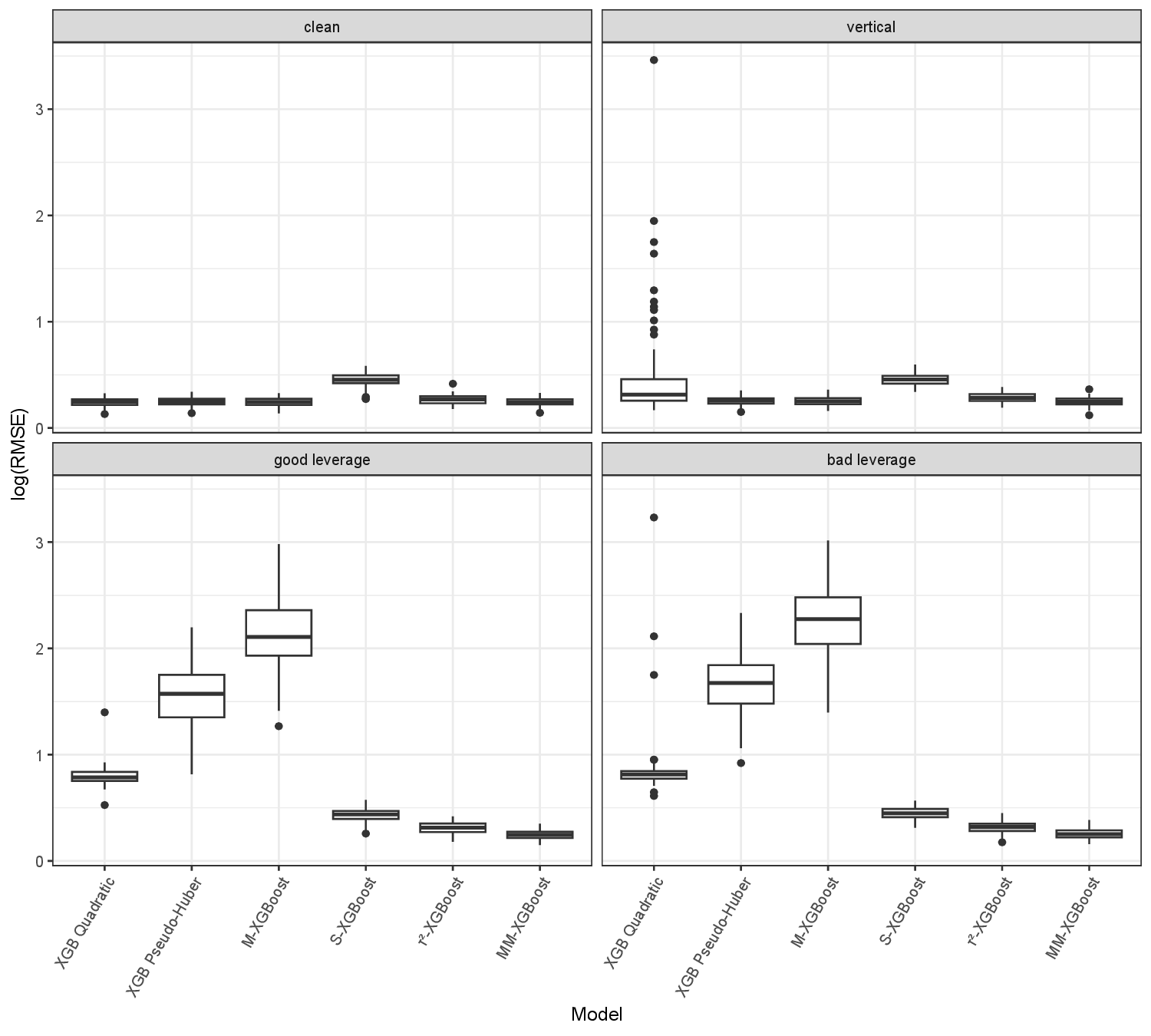}
		\caption{\textit{Boxplots of  $\log({RMSE})$ across 100 independent runs for simulation setting~$1$, and for different contamination scenarios.}}
		\label{fig: boxplot g2, S2, 0.05}
	\end{figure}

	\paragraph{Setting 2: Nonlinear function, high-dimensional, correlated predictors, high contamination rate.}
	The second setting represents a more challenging scenario, combining the nonlinear regression function $g_3$, strong predictor correlation $S_2$, high dimensionality ($p=100$), and a high contamination rate of $20\%$. Figure~\ref{fig: boxplot g3, S1, 0.2} shows the corresponding $\log(\text{RMSE})$ boxplots over $100$ simulation replicates, and Table~\ref{tab:g3_S1_0.2_p100 text} reports the mean RMSE values with standard errors.  
	XGB Quadratic is unstable, with large increases in both prediction error and variability under vertical outliers as well as leverage points. XGB Pseudo-Huber improves robustness to vertical outliers, but the method remains sensitive to leverage points. A similar lack of robustness is observed for M-XGBoost, which is affected by leverage points. In contrast, S-XGBoost remains highly robust across all contamination types, maintaining stable RMSE values. However, its robustness comes at the cost of efficiency: under clean data it yields noticeably higher RMSE than the more efficient XGB Quadratic. The $\tau^2$-XGBoost method improves prediction error relative to S-XGBoost, achieving lower RMSE under clean data. However, this comes with a slight reduction in robustness, as its prediction error increases more noticeably under contamination compared to S-XGBoost. Moreover, it remains less efficient than XGB Quadratic in the clean setting. Finally, MM-XGBoost provides the best overall performance. It substantially improves efficiency compared to S-XGBoost while maintaining strong robustness across all contamination scenarios. In particular, it reaches low prediction error under clean data and maintains stable performance in the presence of vertical outliers, leverage points, strong predictor correlation, high dimensionality, and substantial amounts of outliers. These results demonstrate the ability of MM-XGBoost to combine high efficiency with strong robustness in a more challenging setting.
	
	\begin{table}
		\centering
		\caption{\label{tab:g3_S1_0.2_p100 text}Mean RMSE (SE) for $g_3$ with $p=100$ under clean and 20\% contamination scenarios with $S_2$ for $M = 100$ independent runs}
		\centering
		\begin{tabular}[t]{lcccc}
			\toprule
			Model & clean & vertical & good leverage & bad leverage\\
			\midrule
			XGB Quadratic & 1.68 (0.01) & 12.28 (5.85) & 3.58 (0.02) & 4.09 (0.20)\\
			XGB Pseudo-Huber & 2.71 (0.12) & 2.37 (0.09) & 4.90 (0.12) & 4.82 (0.11)\\
			\midrule
			M-XGBoost & 1.71 (0.01) & 2.18 (0.03) & 3.50 (0.02) & 3.44 (0.03)\\
			\midrule
			S-XGBoost & 3.61 (0.04) & 3.57 (0.04) & 2.78 (0.02) & 2.77 (0.02)\\
			\midrule
			$\tau^2$-XGBoost & 2.34 (0.02) & 2.60 (0.02) & 2.63 (0.02) & 2.65 (0.02)\\
			\midrule
			\addlinespace
			MM-XGBoost & 1.86 (0.01) & 1.90 (0.01) & 1.90 (0.01) & 1.88 (0.01)\\
			\bottomrule
		\end{tabular}
	\end{table}
	
	\begin{figure}
		\centering
		\includegraphics[width=\linewidth]{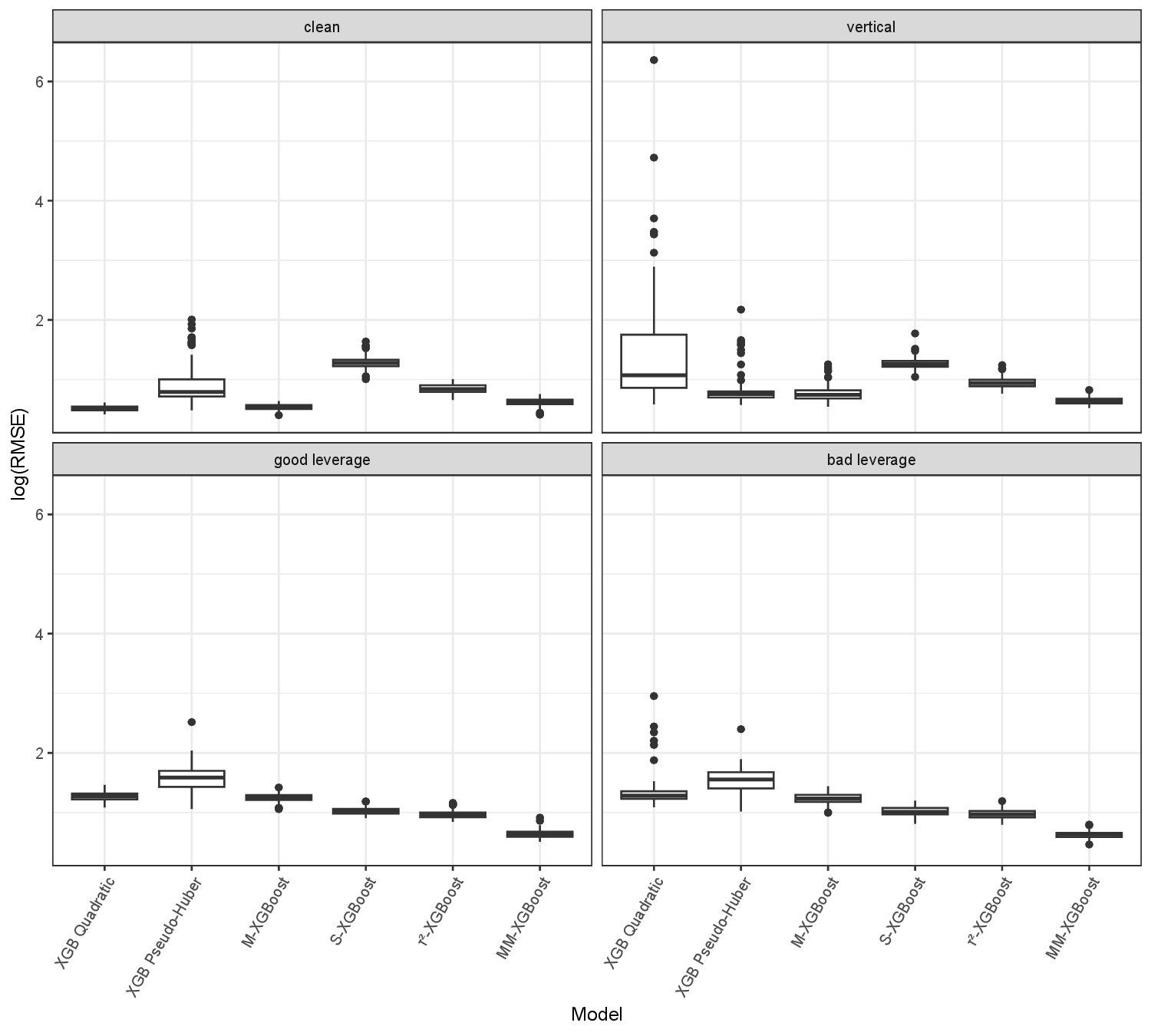}
		\caption{\textit{Boxplots of  $\log({RMSE})$ across 100 independent runs for simulation setting~$2$ and for different contamination scenarios.}}
		\label{fig: boxplot g3, S1, 0.2}
	\end{figure}
	
	\paragraph{Summary of Results Across All Configurations.}
	The general patterns observed in the two representative settings extend across all simulation configurations. From the tables in the supplementary file it is confirmed that  (i) XGB Quadratic is sensitive to all types of contamination, while XGB Pseudo-Huber improves robustness to vertical outliers but remains sensitive to leverage points. M-XGBoost shows a similar lack of robustness to leverage points. 
	(ii) S-XGBoost provides strong robustness in all settings, with stable performance under both vertical outliers and leverage points. However, this robustness is achieved at the cost of increased prediction error. (iii) The $\tau^2$-XGBoost method improves efficiency relative to S-XGBoost, but this gain becomes modest in presence of outliers. (iv) Across all combinations of regression functions, contamination levels, and correlation structures, MM-XGBoost achieves the best overall trade-off between robustness and predictive accuracy. In presence of contamination, it consistently outperforms all competing methods. In absence of outliers, the loss in predictive performance compared to the standard XGBoost is limited.

	\section{Conclusion}\label{sec: conclusion}
	
	This paper proposes  highly robust versions of the widely used XGBoost prediction method for regression. An important practical advantage of the proposed methods is that they are straightforward to implement within the XGBoost framework, which supports customized loss functions. The user only needs to specify the corresponding gradient and diagonal Hessian of the loss function, which have been mathematically derived in Section \ref{sec: methodology}.
	
	The proposed methods were evaluated through extensive simulation studies encompassing a wide range of outlier scenarios. In Section \ref{sec: sim studies} we discussed two representative simulation settings in detail.  Results for other settings are included in the supplementary file, including a situation where the number of predictors $p=400$ exceeds the size of the training set $n=300$. The results demonstrate that standard XGBoost is not robust to outliers, even when employing its built-in Pseudo-Huber loss function. In contrast, the proposed MM-XGBoost method achieves both a high degree of robustness and strong predictive accuracy, making it the most attractive approach overall. 
	
	Our work is closely related to the robust gradient boosting methods proposed by \cite{robustboost}. Specifically, their methods, SBoost and RRBoost, can be viewed as counterparts of S-XGBoost and MM-XGBoost within the standard gradient boosting framework. Unlike XGBoost, standard gradient boosting relies on a first-order  approximation of the loss function and does not require second-order derivative information. A key advantage of XGBoost is its computational efficiency and scalability, which largely explain its widespread popularity. The robust methods proposed in this paper retain these advantages, as they modify only the loss function while preserving the core XGBoost algorithm. Consequently, in our simulation settings, SBoost and RRBoost required up to one hundred times more computation time than S-XGBoost and MM-XGBoost, respectively. In terms of predictive performance, however, the two families of methods yielded broadly comparable results, with MM-XGBoost tending to slightly outperform RRBoost. Hence, MM-XGBoost offers a clear practical advantage by combining the robustness and predictive performance of RRBoost with the computational efficiency and scalability of the XGBoost framework.

	The robust loss functions considered in this paper can also be incorporated into the LightGBM framework \citep{lightgbm}. LightGBM is closely related to XGBoost and is often more computationally efficient, particularly on large datasets. Our exploratory analyses indicate, however, that LightGBM is considerably more vulnerable to outliers than XGBoost, even when robust loss functions are employed. This suggests that robustness may require modifications beyond the loss function alone, potentially involving the histogram-based splitting mechanism. We leave the development of robust LightGBM methods for future research.

	In conclusion, we proposed a robust extension of the widely used XGBoost regression method. By providing an appropriate loss function, the proposed methods retain the computational efficiency and scalability of XGBoost while substantially improving robustness to outliers.  We hope that this work will stimulate further research on robust machine learning methods.
	\newpage
	
	\appendix
	\renewcommand{\thesection}{Appendix}
	\section{Proofs}
	\label{appen:proofs}
	
	\subsection*{Proof of Result~\ref{result: grad and hess M}} \label{appen: sec: proof m-loss}
	\noindent 
	The partial derivative of the M-loss defined in Equation~\eqref{eq: M-XGBoost loss} with respect to $\hat y_j$ is given by
	\begin{equation*}
		g_j
		= \sum_{i=1}^n 
		\psi\big(\frac{r_i}{\sigma_0}\big)
		\big(\frac{-\delta_{ij}}{\sigma_0}\big)\sigma_0^2,
	\end{equation*}
	where $\delta_{ij}$ is the Kronecker delta. Simplifying the expression gives the gradient entry $g_j$ of Equation~\eqref{eq:grad m-xg} in Result~\ref{result: grad and hess M}. Differentiating the gradient entry with respect to $\hat y_j$ once more, yields the diagonal Hessian entry $h_j$ of Equation~\eqref{eq:grad m-xg} in Result~\ref{result: grad and hess M}.
	
	\subsection*{Proof of Result~\ref{result: grad and hess S}}\label{appen: sec: proof s-loss}
	\noindent 
	The first derivative of the S-loss defined in Equation~\eqref{eq: s-loss} with respect to $\hat y_j$, for $j = 1, \dots, n$, is
	\begin{equation}\label{eq:deriv s-loss}
		g_j
		= n \sigma \frac{\partial \sigma}{\partial \hat y_j}.
	\end{equation}
	The M-scale estimator $\sigma$ 
	is given by the solution of the implicit equation
	\begin{equation}\label{eq: m-scale}
		\frac{1}{n}\sum_{i=1}^n \rho_1\big(\frac{r_i}{\sigma}\big) = \kappa,
	\end{equation}
	Define $\psi_1=\rho_1'$ and $\tilde{r}_i = {r_i}/{\sigma}$.
	Differentiating \eqref{eq: m-scale} with respect to $\hat y_j$, for $j = 1, \dots, n$, and setting it equal to zero gives
	\begin{equation} \label{appen: diff m-scale}
		0 = \frac{\partial}{\partial\hat y_j}\big[\frac{1}{n}\sum_{i=1}^n \rho_1(\tilde r_i)\big]
		= \frac{1}{n}\sum_{i=1}^n \psi_1(\tilde r_i)\frac{\partial\tilde r_i}{\partial\hat y_j}.
	\end{equation} 
	The derivative of $\tilde{r}_i$ with respect to $\hat{y}_j$ is
	\begin{equation}\label{appen: deriv tilde r}
		\frac{\partial\tilde r_i}{\partial\hat y_j}
		= \frac{\partial}{\partial\hat y_j}\big(\frac{r_i}{\sigma}\big)
		= \frac{-\delta_{ij}\sigma - r_i \frac{\partial\sigma}{\partial\hat y_j}}{\sigma^2}
		= -\frac{\delta_{ij}}{\sigma} - \frac{\tilde r_i}{\sigma}\frac{\partial\sigma}{\partial\hat y_j}.
	\end{equation}
	Substituting this derivative into \eqref{appen: diff m-scale} and multiplying it by $n\sigma$ results in
	\[
	0 = -\psi_1(\tilde r_j) - \frac{\partial\sigma}{\partial\hat y_j}\sum_{i=1}^n \tilde r_i \psi_1(\tilde r_i).
	\]
	By defining
	\begin{equation}\label{eq:A1-def}
		A_1 \;=\; \sum_{i=1}^n \tilde r_i \psi_1(\tilde r_i),
	\end{equation}
	the first derivative of $\sigma$ with respect to $\hat{y}_j$, for $j = 1, \dots, n$, is now given by,
	\begin{equation}\label{eq:d sigma}
		{\quad
			\frac{\partial\sigma}{\partial\hat y_j} \;=\; -\frac{\psi_1(\tilde r_j)}{A_1}.
			\quad}
	\end{equation}
	Substituting \eqref{eq:d sigma} into \eqref{eq:deriv s-loss} yields the gradient entry~\eqref{eq:gradS} in Result~\ref{result: grad and hess S}.
	The diagonal Hessian entry is obtained by differentiating the gradient entry with respect to $\hat y_j$ once more:
	\begin{equation}\label{eq: second deriv S-loss}
		h_j
		= n\big(
		\sigma\,\frac{\partial^2 \sigma}{\partial \hat y_j^2}
		+ \Big(\frac{\partial \sigma}{\partial \hat y_j}\Big)^2
		\big).
	\end{equation}
	The second derivative of the M-scale estimator $\sigma$, is obtained by differentiating the identity of the first derivative \eqref{appen: diff m-scale} again with respect to $\hat{y}_j$ for $j = 1, \dots, n$:
	\begin{equation} \label{appen eq: second deriv sigma}
		\frac{\partial}{\partial\hat y_j}\psi_1(\tilde r_j)
		+ \frac{\partial^2\sigma}{\partial\hat y_j^2}\, A_1
		+ \frac{\partial\sigma}{\partial\hat y_j}\frac{\partial A_1}{\partial\hat y_j}
		= 0.
	\end{equation}
	We evaluate the first and third term of above equation.
	For the first term, using the chain rule we obtain
	\[
	\frac{\partial}{\partial \hat y_j}\psi_1(\tilde r_j) = \psi_1'(\tilde r_j)\frac{\partial\tilde r_j}{\partial\hat y_j}
	= \psi_1'(\tilde r_j)\big(-\frac{1}{\sigma} - \frac{\tilde r_j}{\sigma}\frac{\partial\sigma}{\partial\hat y_j}\big).
	\]
	For the third term, we compute ${\partial A_1}/{\partial \hat y_j}$, 
	\begin{align*} 
		\frac{ \partial A_1}{\partial\hat y_j}
		= \sum_{i=1}^n \frac{\partial}{\partial \hat y_j}\big(\tilde r_i \psi_1(\tilde r_i)\big)
		&= \sum_{i=1}^n \big\{\frac{\partial\tilde r_i}{\partial\hat y_j}\psi_1(\tilde r_i)
		+ \tilde r_i \psi_1'(\tilde r_i)\frac{\partial\tilde r_i}{\partial\hat y_j}\big\} \\ &= \sum_{i=1}^n \frac{\partial\tilde r_i}{\partial\hat y_j}\big(\psi_1(\tilde r_i) + \tilde r_i \psi_1'(\tilde r_i)\big).
	\end{align*}
	Define \begin{equation}\label{eq:B1-def}
		B_1 \;=\; \sum_{i=1}^n \tilde r_i\big(\psi_1(\tilde r_i) + \tilde r_i\psi_1'(\tilde r_i)\big).
	\end{equation} 
	Substituting the derivative of $\tilde r_i$ \eqref{appen: deriv tilde r} into above equation yields 
	\[
	\frac{\partial A_1}{\partial \hat y_j}
	= -\frac{1}{\sigma}\big(\psi_1(\tilde r_j) + \tilde r_j\psi_1'(\tilde r_j)\big)
	- \frac{\partial\sigma}{\partial\hat y_j}\frac{B_1}{\sigma}.
	\]
	Inserting the evaluated first and third terms into~\eqref{appen eq: second deriv sigma} leads to
	\[
	\psi_1'(\tilde r_j)\Big(-\frac{1}{\sigma} - \frac{\tilde r_j}{\sigma}\frac{\partial\sigma}{\partial\hat y_j}\Big)
	+ \frac{\partial^2\sigma}{\partial\hat y_j^2}\,A_1
	+ \frac{\partial\sigma}{\partial\hat y_j}\Big[-\frac{1}{\sigma}\big(\psi_1(\tilde r_j) + \tilde r_j\psi_1'(\tilde r_j)\big)
	- \frac{\partial\sigma}{\partial\hat y_j}\frac{B_1}{\sigma}\Big] = 0.
	\]
	Simplifying the above equation and writing in terms of  $\partial^2\sigma/\partial\hat y_j^2$ gives
	\[
	\sigma A_1 \frac{\partial^2\sigma}{\partial\hat y_j^2}
	= \psi_1'(\tilde r_j) + \psi_1'(\tilde r_j)\tilde r_j\frac{\partial\sigma}{\partial\hat y_j}
	+ \frac{\partial\sigma}{\partial\hat y_j}\big(\psi_1(\tilde r_j) + \tilde r_j\psi_1'(\tilde r_j)\big)
	+ \frac{B_1}{\sigma}\big(\frac{\partial\sigma}{\partial\hat y_j}\big)^2.
	\]
	Finally, we substitute $\partial\sigma/\partial\hat y_j = -\psi_1(\tilde r_j)/A_1$, obtained in \eqref{eq:d sigma}, in the above equation. After reordering the terms, the second derivative of $\sigma$ with respect to $\hat y_j$, for $j=1, \dots, n$, is
	\begin{equation}\label{eq:d2sigma}
		{\quad
			\frac{\partial^2\sigma}{\partial\hat y_j^2}
			= \frac{\psi_1'(\tilde r_j)}{\sigma A_1}
			- 2\,\frac{\psi_1'(\tilde r_j)\psi_1(\tilde r_j)\tilde r_j}{\sigma A_1^2}
			+ \frac{\psi_1^2(\tilde r_j)}{\sigma A_1^2}\big(\frac{B_1}{A_1}-1\big)
			\quad}
	\end{equation}
	where $A_1$ and $B_1$ are defined in \eqref{eq:A1-def} and \eqref{eq:B1-def} respectively. 
	Substituting \eqref{eq:d sigma} and \eqref{eq:d2sigma} into \eqref{eq: second deriv S-loss} yields the diagonal Hessian entry~\eqref{eq:hessS} in Result~\ref{result: grad and hess S}.
	
	\subsection*{Proof of Result~\ref{result: grad and hess tau}}\label{appen: sec: proof tau-loss}
	\noindent
	The first derivative of the $\tau^2$-loss derived in Equation~\eqref{eq: tau2 loss} with respect to $\hat{y}_j$, for $j = 1, \dots, n$, follows directly from the chain rule. Since  the $\tau^2$-loss depends on $\hat{y}_j$ through $\tau^2$, we obtain
	\begin{equation}\label{eq: grad tau-loss}
		g_j
		= \frac{n}{2\kappa}\,\frac{\partial\tau^2}{\partial\hat y_j}.
	\end{equation}
	The $\tau$-scale estimator is given by
	\begin{equation}\label{appen: tau def}
		\tau^2 = \frac{1}{n}\sum_{i=1}^n \rho\big(\frac{r_i}{\sigma}\big)\sigma^2,
	\end{equation}
	with $\sigma$ an M-scale estimator which is the solution to the equation
	\begin{equation*}
		\frac{1}{n}\sum_{i=1}^n \rho_1\big(\frac{r_i}{\sigma}\big) = \kappa.
	\end{equation*}
	To obtain the derivative of the $\tau$-scale estimator we differentiate \eqref{appen: tau def} with respect to $\hat y_j$ for $j = 1,\dots n$. Note that $\tau^2$ depends on $\hat{y}_j$ both directly and through $\sigma$. This results in
	\begin{equation}\label{appen: deriv tau}
		\frac{\partial}{\partial \hat y_j}\big[\frac{1}{n}\sum_{i=1}^n \rho(\tilde r_i)\sigma^2\big]
		= \frac{1}{n}\sum_{i=1}^n \psi(\tilde r_i)\frac{\partial\tilde r_i}{\partial\hat y_j}\sigma^2 + \frac{1}{n}\sum_{i=1}^n \rho(\tilde r_i)2\sigma \frac{\partial\sigma}{\partial\hat y_j} ,
	\end{equation}
	where $\psi=\rho'$ and $\tilde{r}_i = {r_i}/{\sigma}$. 
	Define
	\begin{equation}\label{eq: C and A def}
		C = \sum_{i=1}^n\rho(\tilde r_i), \qquad \text{and} \qquad A = \sum_{i=1}^n \tilde r_i \psi(\tilde r_i).
	\end{equation}
	Substituting the derivative of $\tilde r_i$ \eqref{appen: deriv tilde r} and the derivative of $\sigma$ \eqref{eq:d sigma} into~\eqref{appen: deriv tau} gives
	\begin{equation} \label{appen: eq: deriv tau2 step}
		\frac{\partial \tau^2}{\partial \hat y_j}
		= -\frac{\psi_1(\tilde{r}_j)}{A_1}
		\big(\frac{-1}{n}A\sigma + 2\sigma\frac{1}{n}C \big)
		- \frac{\psi(\tilde{r}_j)\sigma}{n},
	\end{equation}
	with 
	$A_1$ defined in \eqref{eq:A1-def}.
	After reordering Equation~\eqref{appen: eq: deriv tau2 step} we get the first derivative of the $\tau$-scale estimator with respect to $\hat{y}_j$, for $j=1,\dots,n$,
	\begin{equation}\label{eq:tau2-first-deriv}
		{\quad
			\frac{\partial \tau^2}{\partial \hat y_j}
			= -\frac{\psi_1(\tilde{r}_j)\sigma}{n}
			\big( \frac{2C}{A_1} - \frac{A}{A_1} \big)
			- \frac{\psi(\tilde{r}_j)\sigma}{n}.}
	\end{equation}
	Substituting \eqref{eq:tau2-first-deriv} into \eqref{eq: grad tau-loss} yields the gradient entry~\eqref{eq:grad-tau2} in Result~\ref{result: grad and hess tau}. The diagonal Hessian entry is obtained by differentiating the gradient entry with respect to $\hat y_j$ once more,
	\begin{equation*}
		h_j
		= \frac{n}{2\kappa} \frac{\partial^2 \tau^2}{\partial\hat y_j^2}.
	\end{equation*}
	To obtain the second derivative of the $\tau$-scale estimator $\tau^2$, we differentiate the identity of the first derivative \eqref{appen: deriv tau} once again with respect to $\hat{y}_j$ for $j = 1, \dots, n$,
	\begin{align} \label{eq: form second deriv tau}
		\frac{\partial^2 \tau^2}{\partial\hat y_j^2} 
		= \frac{1}{n}\big[
		\frac{\partial}{\partial\hat y_j} \big(-\sigma A \frac{\partial \sigma}{\partial \hat y_j}\big) 
		+ \frac{\partial}{\partial\hat y_j}\big(2\sigma C\frac{\partial \sigma}{\partial \hat y_j}\big)
		- \frac{\partial}{\partial\hat y_j}(\sigma \psi(\tilde r_j)) \big].
	\end{align}
	We evaluate the terms of the above equation separately. Define \begin{equation*}\label{eq:B-def}
		B \;=\; \sum_{i=1}^n \tilde r_i\big(\psi(\tilde r_i) + \tilde r_i\psi'(\tilde r_i)\big).
	\end{equation*}
	Expanding and simplifying the first term in~\eqref{eq: form second deriv tau} gives
	\begin{align*}
		\frac{\partial}{\partial\hat y_j} \big(-\sigma A \frac{\partial \sigma}{\partial \hat y_j}\big) &= -A\big(\frac{\partial\sigma}{\partial\hat y_j}\big)
		^2 -\sigma\big(\frac{-v'(\tilde r_j)}{\sigma}-\frac{1}{\sigma} \frac{\partial \sigma}{\partial \hat y_j}B\big)\frac{\partial \sigma}{\partial \hat y_j} - \sigma A \frac{\partial^2\sigma}{\partial\hat y_j^2} \\
		&= \big(\frac{\partial \sigma}{\partial \hat y_j}\big)^2 (B-A) - \sigma A \frac{\partial^2 \sigma}{\partial \hat y_j^2} + v'(\tilde r_j) \frac{\partial \sigma}{\partial \hat y_j}.
	\end{align*}
	Expanding the second term in~\eqref{eq: form second deriv tau} gives the equation
	\begin{align}\label{appen: term 2 second deriv tau}
		\frac{\partial}{\partial\hat y_j}\big(2\sigma C\frac{\partial \sigma}{\partial \hat y_j}\big) = 2 C \big(\frac{\partial \sigma}{\partial \hat y_j}\big)^2 + 2\sigma \big(\frac{\partial C}{\partial\hat y_j} \frac{\partial \sigma}{\partial \hat y_j} + C \frac{\partial^2 \sigma}{\partial \hat y_j^2}\big).
	\end{align}
	To evaluate this second term we compute ${\partial C}/{\partial \hat y_j}$, with $C$ defined in~\eqref{eq: C and A def}. Using the derivative of $\tilde r_i$ obtained in~\eqref{appen: deriv tilde r} we get
	\begin{align*}
		\frac{\partial C}{\partial \hat y_j} = \frac{\partial}{\partial \hat y_j} \sum_{i=1}^n \rho(\tilde r_i) = \sum_{i=1}^n \psi(\tilde r_i) \frac{\partial \tilde r_i}{\partial \hat y_j} = \frac{-A}{\sigma} \frac{\partial \sigma}{\partial \hat y_j} - \frac{\psi(\tilde r_j)}{\sigma}.
	\end{align*}
	Plugging the derivative of $C$ into the second term \eqref{appen: term 2 second deriv tau} and rearranging the terms gives now the second term
	\begin{align*}
		\frac{\partial}{\partial\hat y_j}\big(2\sigma C\frac{\partial \sigma}{\partial \hat y_j}\big) = 2(C-A) \big(\frac{\partial \sigma}{\partial \hat y_j}\big)^2 + 2 \sigma C \frac{\partial^2 \sigma}{\partial \hat y_j^2} - 2 \psi(\tilde r_j) \frac{\partial \sigma}{\partial \hat y_j}.
	\end{align*}
	The third term of the second derivative \eqref{eq: form second deriv tau} is 
	\begin{align*}
		\frac{\partial}{\partial\hat y_j}\big(-\sigma \psi(\tilde r_j)\big)= -\psi(\tilde r_j) \frac{\partial \sigma}{\partial \hat y_j} - \sigma \psi'(\tilde r_j) \frac{\partial \tilde r_j}{\partial \hat y_j}.
	\end{align*}
	Substituting the derivative of $\tilde r_i$ obtained in \eqref{appen: deriv tilde r} gives
	\begin{align*}
		\frac{\partial}{\partial\hat y_j}(-\sigma \psi(\tilde r_j))= \frac{\partial \sigma}{\partial \hat y_j}\big(\psi'(\tilde r_j)\tilde r_j-\psi(\tilde r_j) \big) - \psi'(\tilde r_j).
	\end{align*}
	Finally, we insert all evaluated terms into the second derivative equation of the $\tau$-scale estimator \eqref{eq: form second deriv tau} w.r.t $\hat y_j$ for $j = 1, \dots, n$, which results in
	\begin{align}
		\frac{\partial^2 \tau^2}{\partial\hat y_j^2} 
		= \frac{1}{n}\Big[
		&\big(\frac{\partial\sigma}{\partial\hat y_j}\big)^2 (2C + B - 3A) \nonumber\\
		+ &\big(\frac{\partial^2\sigma}{\partial\hat y_j^2}\big)\big(2\sigma C - \sigma A\big) \nonumber\\
		+ &\big(\frac{\partial\sigma}{\partial\hat y_j}\big)\big(2 \psi'(\tilde{r}_j)\tilde{r}_j - 2\psi(\tilde{r}_j)\big)
		+ \psi'(\tilde{r}_j)
		\Big].
	\end{align}
	Substituting \eqref{eq:d sigma} and \eqref{eq:d2sigma} into the second derivative of $\tau^2$ yields the diagonal Hessian entry~\eqref{eq:final_Hessian_tau2} in Result~\ref{result: grad and hess tau}.

	\newpage
	\bibliographystyle{apalike}
	\bibliography{biblio}
\end{document}